\documentclass[conference]{IEEEtran}
\IEEEoverridecommandlockouts

\usepackage{paper}

\def\BibTeX{{\rm B\kern-.05em{\sc i\kern-.025em b}\kern-.08em
    T\kern-.1667em\lower.7ex\hbox{E}\kern-.125emX}}

\makeatletter
\def\ps@IEEEtitlepagestyle{%
  \def\@oddfoot{\mycopyrightnotice}%
}
\def\mycopyrightnotice{%
\fbox{\parbox{\dimexpr\textwidth-2\fboxsep-2\fboxrule\relax}{
\begin{minipage}{\textwidth-2\fboxsep-2\fboxrule}
  \footnotesize
  \textcopyright 2024 IEEE. Personal use of this material is permitted. Permission from IEEE must be obtained for all other uses, in any current or future media, including reprinting/republishing this material for advertising or promotional purposes, creating new collective works, for resale or redistribution to servers or lists, or reuse of any copyrighted component of this work in other works.
  \end{minipage}
}}
}
\makeatother

\begin{document}


\title{Advancing Risk and Quality Assurance: A RAG Chatbot for Improved Regulatory Compliance}

\author{\IEEEauthorblockN{
Lars Hillebrand\IEEEauthorrefmark{1}\IEEEauthorrefmark{2}\IEEEauthorrefmark{5}\thanks{* Both authors contributed equally to this research.}, Armin Berger\IEEEauthorrefmark{1}\IEEEauthorrefmark{2}\IEEEauthorrefmark{3}, Daniel Uedelhoven\IEEEauthorrefmark{2}\IEEEauthorrefmark{5}, David Berghaus\IEEEauthorrefmark{2}\IEEEauthorrefmark{5},\\Ulrich Warning\IEEEauthorrefmark{4}, Tim Dilmaghani\IEEEauthorrefmark{4}, Bernd Kliem\IEEEauthorrefmark{4}, Thomas Schmid\IEEEauthorrefmark{4},
Rüdiger Loitz\IEEEauthorrefmark{4}, 
Rafet Sifa\IEEEauthorrefmark{2}\IEEEauthorrefmark{3}} \\
\IEEEauthorrefmark{2}\textit{Fraunhofer IAIS}, Sankt Augustin, Germany \\
\IEEEauthorrefmark{3}\textit{University of Bonn}, Bonn, Germany\\
\IEEEauthorrefmark{4}\textit{PricewaterhouseCoopers GmbH}, Düsseldorf, Germany\\
\IEEEauthorrefmark{5}\textit{Lamarr Institute}, Germany
}
\IEEEpubid{\makebox[\columnwidth]{979-8-3503-2445-7/23/\$31.00 ©2023 IEEE \hfill} \hspace{\columnsep}\makebox[\columnwidth]{ }}

\maketitle

\begin{abstract}

Risk and Quality (R\&Q) assurance in highly regulated industries requires constant navigation of complex regulatory frameworks, with employees handling numerous daily queries demanding accurate policy interpretation. Traditional methods relying on specialized experts create operational bottlenecks and limit scalability. We present a novel Retrieval Augmented Generation (RAG) system leveraging Large Language Models (LLMs), hybrid search and relevance boosting to enhance R\&Q query processing. Evaluated on 124 expert-annotated real-world queries, our actively deployed system demonstrates substantial improvements over traditional RAG approaches. Additionally, we perform an extensive hyperparameter analysis to compare and evaluate multiple configuration setups, delivering valuable insights to practitioners.


\end{abstract}

\begin{IEEEkeywords}
Large Language Models, Retrieval Augmented Generation, Legal, Compliance
\end{IEEEkeywords}

\section{Introduction}




Compliance with Risk Management \& Quality Standards is fundamental in regulated industries like auditing, finance, and legal services, where non-compliance can lead to significant legal penalties and financial losses. 
Employees face the challenge of navigating complex regulations and policies, handling numerous internal queries daily that require nuanced interpretation of trusted sources.

\begin{figure*}[ht]
    \centering
    \includegraphics[width=0.70\linewidth]{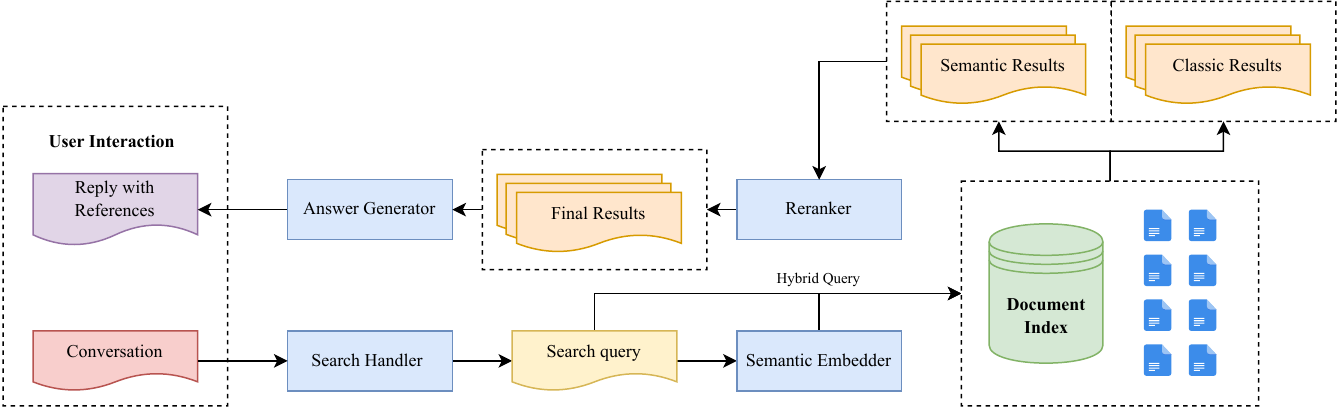}
    \caption{Architecture of the Retrieval Augmented Generation chatbot system, demonstrating the workflow for query resolution.}
    \label{fig:rq_system}
\end{figure*}

Large Language Models (LLMs) such as GPT \cite{brown2020language, OpenAI23} are becoming increasingly important in legal work \cite{rodgers2023a}. We propose a specialized ChatBot powered by GPT with an optimized Retrieval Augmented Generation (RAG) pipeline. Our system (see Figure \ref{fig:rq_system}) integrates hybrid search and relevance boosting to enhance retrieval accuracy and response quality.

We evaluate our solution using a handcrafted dataset with expert-annotated answers and develop a custom evaluation framework using \texttt{DeepEval}\footnote{\url{https://github.com/confident-ai/deepeval}.}. The framework achieves a $0.70$ correlation coefficient with expert assessments. The ChatBot has been successfully deployed within the R\&Q department of PricewaterhouseCoopers GmbH. The key contributions of this work are:

\begin{figure}[t]
    \centering
    \begin{tcolorbox}[colback=gray!5,colframe=black!50,title={Example Query and R\&Q Chatbot Response}]
\scriptsize
\textbf{Employee query:}\\
\textit{``How far does the prohibition on the provision of tax advisory services extend in the case of EU PIEs?''}\\

\textbf{Workflow:}\\
\texttt{Preparing and embedding query}\\
\texttt{Running retrieval using hybrid search}\\
\texttt{Re-ranking text- and vector search results}\\
\texttt{Generating final answer with references}\\

\textbf{Final answer:}\\
\textit{``To answer the question of how far the prohibition on providing tax advisory services to EU Public Interest Entities (PIEs) extends, we need to analyze the relevant information from the provided documents. [...] In summary, the prohibition on providing tax advisory services to EU-PIEs in Germany is very extensive and encompasses almost all tax advisory services listed on the blacklist. However, there are specific exceptions and national regulations that may allow certain tax advisory services under certain conditions.''}
\end{tcolorbox}
    \caption{Example user question, model workflow and generated answer.}
    \label{fig:example_query}
\end{figure}




\begin{itemize}
    \item \textbf{Development of a RAG ChatBot for R\&Q standards:} We introduce a specialized ChatBot combining advanced AI capabilities with RAG.
    \item \textbf{Establishment of a Robust Evaluation Framework:} We devise an automated chatbot evaluation method corroborated by expert assessments.
    \item \textbf{Insights into Hyperparameter Optimization:} We identify how core hyperparameters affect system performance.
\end{itemize}

\section{Methodology} \label{section:methodology}

This section presents our system designed to enhance compliance with Risk Management \& Quality standards through ML-driven solutions. The system comprises: (1) an ingestion pipeline for document processing and indexing; (2) a RAG chatbot leveraging LLMs; and (3) an automated evaluation framework.

\subsection{Ingestion Pipeline and Knowledge Base Construction}



Our ingestion pipeline uses \texttt{Unstructured}\footnote{\url{https://github.com/Unstructured-IO/unstructured}.} to parse documents into a structured data model. Documents are chunked with overlap for context continuity, with internal document chunks receiving a $2\times$ boosting factor. Embeddings are generated using OpenAI's \texttt{ada-002} and \texttt{3-large} \cite{text-embedding-3-large} endpoints and indexed via Azure AI Search.

\subsection{Retrieval Augmented Generation Chatbot}

Our RAG chatbot system interprets user queries, retrieves relevant information from the knowledge base, and generates contextually appropriate responses. We employ a hybrid search strategy that combines vector similarity search and full-text search using TF-IDF-based BM25 algorithms. The results from both searches are re-ranked using reciprocal rank fusion to enhance retrieval effectiveness. Moreover, we utilize the above described relevance boosting to ensure that the chatbot provides answers based on the most trusted and relevant information.

\subsection{Automated Evaluation Framework}

We establish an automated evaluation framework using \texttt{DeepEval}\footnote{\url{https://github.com/confident-ai/deepeval}.} and the G-Eval scoring method~\cite{liu2023g}. The evaluation focuses on correctness, completeness, relevance, and adherence to R\&Q standards. We leverage GPT-4o as the LLM backbone for the evaluation and define the metric range between 0 (worst) and 5 (best). The following evaluation steps are performed to create the final score per sample.
\begin{tcolorbox}[breakable,colback=gray!5,colframe=black!50,title={Evaluation Steps}]
\scriptsize
\textbf{Answer Evaluation Steps}
\begin{itemize}[leftmargin=*]
    \item Check if the facts in 'Actual Output' contradict any facts in 'Expected Output'.
    \item DO NOT punish long and detailed answers if the 'Actual Output' is perfectly correct. Generally, more details in the 'Actual Output' are encouraged.
    \item If the 'Actual Output' misses details compared to the 'Expected Output' you should slightly penalize omission of detail.
\end{itemize}

\textbf{Context Evaluation Steps}
\begin{itemize}[leftmargin=*]
    \item Summarize the expected 'Context' and note the most important points.
    \item Compare the summary with the 'Retrieval Context' and check if the most important points are present.
    \item If the 'Retrieval Context' is missing important points compared to the 'Context' you should penalize the response.
    \item If the 'Retrieval Context' contains irrelevant information, you should very slightly penalize the response.
    \item If the 'Retrieval Context' contains contradictory information, you should heavily penalize the response.
\end{itemize}
\end{tcolorbox}
To validate reliability, we compare LLM-based scores with manual evaluations from domain experts across 124 responses, achieving a Pearson correlation coefficient of $r = 0.70$. While acknowledging potential LLM biases~\cite{zheng2023judging}, this correlation supports the use of automated evaluations as proxies for expert judgment.

\section{Experiments} \label{section:experiments}

We conduct experiments on an expert-curated dataset to provide insights for implementing LLM-based chatbots in production environments. We present our dataset, experimental setup, and discuss our findings.

\subsection{Data}
Our dataset\footnote{Dataset and Python code are currently unpublishable due to ongoing industrial project constraints.} comprises 124 R\&Q question-answer pairs created by domain experts. 
An illustrative question example is highlighted in Figure \ref{fig:example_query}. Of these questions, 110 used internal sources, with the remainder drawing from external data. 
Thirteen experts contributed to the dataset creation, with oversight from three senior R\&Q specialists. Multiple review rounds ensured quality control through random sampling and qualitative assessment.

\subsection{Model Configurations}

Our ablation studies examine three key areas: (1) ingestion parameters, (2) retrieval parameters, and (3) model parameters, measuring their impact on system performance. Table~\ref{tab:ablation_configurations} presents the complete configuration space, with bold values indicating our baseline setup. All configurations use an LLM temperature value of $0$ to increase answer robustness. Through systematic evaluation of ingestion and retrieval parameters, we identify the optimal configuration achieving the highest correctness scores. While initially using \texttt{ada-002} for embeddings, we discovered that \texttt{3-large} yields superior performance during our retrieval optimization process, leading to its adoption in subsequent experiments. The final optimized configuration is then used as the foundation to assess different LLM backbones (see Table \ref{tab:model_performance_metrics}).
\begin{table}[t]
\centering
\caption{Hyperparameter configurations. Bold values indicate the Baseline setup used for ablation studies.}
\begin{tabular}{llc}
\toprule
Module & Hyperparameter &  Configurations \\
\midrule
\multirow{4}{*}{Ingestion}  & Max Chunk Size                & $256$, $\bf{512}$, $1024$, $2048$ \\
                            & Min Chunk Overlap             & $32$, $\bf{64}$, $128$, $256$  \\
                            & Markdown Conversion           & Yes, \textbf{No} \\
\midrule
\multirow{4}{*}{Search}  & Top-k                 & $5$, $\bf{10}$, $20$ \\
                            & Search Type           & Text, \textbf{Hybrid}, Vector \\
                            & Relevance boosting    & Yes, \textbf{No} \\
                            & Embedding model    & ada-002, 3-large \\

\midrule
\multirow{1}{*}{ChatBot}  & LLM-Backbone (GPT)          & 4o-mini, \textbf{4o}, 3.5-Turbo, 4-Turbo \\
\bottomrule
\end{tabular}
\label{tab:ablation_configurations}
\end{table}

\subsection{Prompt Design}
\label{subsec:prompt_design}

Our prompt design includes a template that ensures consistent and accurate responses from the LLM. We utilize dynamic language detection using the \texttt{langdetect}\footnote{\url{https://github.com/Mimino666/langdetect}.} library to automatically adjust the language of the response to match the user's query. The prompt instructs the model to cite sources appropriately, and avoid hallucinations by stating when information is not present in the provided context.
\begin{tcolorbox}[colback=gray!5,colframe=black!50,title=Prompt Template]
\scriptsize
\{\texttt{user\_question}\}\\

\textless instruction\textgreater

Write your answer in \{\texttt{language}\}. If you cannot answer the question based on the provided context, state that the information is not present, don't invent or hallucinate an answer and don't reference any sources. After each fact you state, provide the corresponding document name and chunk id from the appended sources in brackets and separated by "/". For example: "Apple was founded in 1976." \textasteriskcentered(apple.docx/1)\textasteriskcentered 

Don't combine sources but list each individual source separately if a fact contains multiple sources. E.g. \textasteriskcentered(apple.docx/1)\textasteriskcentered, \textasteriskcentered(apple.docx/2)\textasteriskcentered, etc. 

You must comply with the following sources format: \textasteriskcentered(\textless document\_name\_as\_str\textgreater/\textless chunk\_id\_as\_int\textgreater)\textasteriskcentered 

Before answering the question, lay out your full thought process and dissect the user question and its implications.

\textless /instruction\textgreater \\

\textless document\_context\textgreater

\{\texttt{retrieved\_chunks}\}

\textless /document\_context\textgreater
\end{tcolorbox}

\subsection{Results}







Our experiments reveal that the optimal configuration, Baseline$_{\text{3-large}}$ with relevance boosting enabled, achieves the highest correctness scores for both answers and context, as detailed in Table~\ref{tab:ablation_study}. Hybrid search consistently outperforms individual methods, and relevance boosting further improves the prioritization of internal documents.

Using this optimal configuration, named R\&Q-Chatbot, we conduct a comprehensive evaluation across different LLM backbones. Table~\ref{tab:model_performance_metrics} shows that while all models deliver reasonable answers, GPT-4o demonstrates the best performance. For robust analysis, each model configuration was evaluated using 5 independent runs, reporting mean and standard deviation for our G-Eval correctness metric.

\begin{table}[t]
\centering
\caption{Detailed ablation study to evaluate multiple model configurations. We report mean and standard deviation values of 5 independent runs (best scores in bold) for both, answer and context correctness (Scale: 0-5).}
\begin{tabular}{lrr}
\toprule
\textbf{Model Configuration} & \multicolumn{2}{c}{\textbf{G-Eval Correctness Score}} \\
\cmidrule(r){2-3}
& Answer $\uparrow$& Context $\uparrow$\\
\midrule
Baseline$_\text{ada-002}$ & $\bm{3.72}$ ($\pm.026$)& $2.80$ ($\pm.028$)\\
\quad Chunking: 256/64 & $3.61$ ($\pm.048$)& $2.75$ ($\pm.041$)\\
\quad Chunking: 512/32 & $3.69$ ($\pm.042$)& $2.84$ ($\pm.043$)\\
\quad Chunking: 512/128 & $3.69$ ($\pm.053$) & $\bm{2.88}$ ($\pm.046$)\\
\quad Chunking: 1024/128 & $3.67$ ($\pm.045$)& $2.74$ ($\pm.049$)\\
\quad Chunking: 1024/256 & $3.66$ ($\pm.039$)& $2.83$ ($\pm.045$)\\
\quad Chunking: 2048/256 & $3.56$ ($\pm.050$)& $2.29$ ($\pm.015$)\\
\quad +Markdown & $3.65$ ($\pm.050$)& $2.77$ ($\pm.030$)\\
\midrule
Baseline$_\text{3-large}$ & $3.76$ ($\pm .030$)& $\bm{2.91}$ ($\pm.031$)\\
\quad Vector Search & $3.72$ ($\pm.030$)& $\bm{2.91}$ ($\pm.032$)\\
\quad Text Search & $3.60$ ($\pm.030$)& $2.62$ ($\pm.026$)\\
\quad Top-k: 5 & $3.72$ ($\pm.033$)& $2.77$ ($\pm.027$)\\
\quad Top-k: 20 & $3.72$ ($\pm.016$)& $2.90$ ($\pm.048$)\\
\quad +Relevance Boosting & $\bm{3.79}$ ($\pm.037$)& $2.90$ ($\pm.018$)\\
\bottomrule
\addlinespace[0.5ex]
\multicolumn{3}{l}{\scriptsize{Chunking: 512/64 (Max Chunk Size = 512, Min Chunk Overlap = 64)}}
\end{tabular}
\label{tab:ablation_study}
\end{table}

\begin{table}[t]
\centering
\caption{Results of the best architectural setup for different LLM backbones (Scale: 0-5 and best scores in bold).}
\begin{tabular}{lrr}
\toprule
\textbf{Model Configuration} & \multicolumn{2}{c}{\textbf{G-Eval Correctness Score}} \\
\cmidrule(r){2-3}
& Answer $\uparrow$& Context $\uparrow$\\
\midrule
GPT-4o (R\&Q-Chatbot) & $\bm{3.79}$ ($\pm.037$)& $\bm{2.90}$ ($\pm.018$)\\
GPT-4-Turbo & $3.69$ ($\pm.047$)& $2.84$ ($\pm.048$)\\
GPT-4o-mini & $3.63$ ($\pm.053$)& $2.79$ ($\pm.037$)\\
GPT-3.5-Turbo & $3.27$ ($\pm.012$)& $2.53$ ($\pm.077$)\\

\bottomrule
\end{tabular}
\label{tab:model_performance_metrics}
\end{table}

\section{Conclusion and Future Work}
\label{section:conclusion}

In this work, we introduced a novel RAG chatbot system tailored for R\&Q assurance in highly regulated industries. Our system effectively leverages LLMs with optimized retrieval strategies including hybrid search and relevance boosting to improve query processing and compliance adherence. The evaluation demonstrates significant performance gains over baseline approaches, validating the efficacy of our system.

Future research will focus on extending the chatbot to a dynamic multi-agent system capable of intelligent query dissection, clarifying questions, and multi-hop reasoning to further enhance its conversational capabilities.

\section{Acknowledgment}

This research has been funded by the Federal Ministry of Education and Research of Germany and the state of North-Rhine Westphalia as part of the Lamarr-Institute for Machine Learning and Artificial Intelligence, LAMARR22B.

\renewcommand*{\bibfont}{\footnotesize}
\printbibliography

\end{document}